\documentclass[10pt,twocolumn,letterpaper]{article}

\usepackage{cvpr}
\usepackage{times}
\usepackage{epsfig}
\usepackage{graphicx}
\usepackage{amsmath}
\usepackage{amssymb}
\usepackage{epigraph}
\usepackage{booktabs}
\usepackage{color}
\usepackage{tipa}
\usepackage{caption}
\usepackage{subcaption}
\usepackage{enumitem}
\usepackage{tabularx}
\usepackage{fancyvrb}

\newcolumntype{L}[1]{>{\raggedright\arraybackslash}p{#1}}
\newcolumntype{C}[1]{>{\centering\arraybackslash}p{#1}}
\newcolumntype{R}[1]{>{\raggedleft\arraybackslash}p{#1}}

\usepackage[breaklinks=true,bookmarks=false]{hyperref}

\cvprfinalcopy 

\def\cvprPaperID{****} 

\newcommand{\norm}[1]{\lVert#1\rVert}

\begin{document}

\title{Split-Brain Autoencoders: \\ Unsupervised Learning by Cross-Channel Prediction \vspace{-3mm}}

\author{Richard Zhang\\
\and Phillip Isola\\
\and Alexei A. Efros\\
\and \\
Berkeley AI Research (BAIR) Laboratory \\
University of California, Berkeley \\
{\tt\small \{rich.zhang,isola,efros\}@eecs.berkeley.edu}
\vspace{-4mm}
}

\maketitle

\begin{abstract}

\vspace{-1mm}
We propose split-brain autoencoders, a straightforward modification of the traditional autoencoder architecture, for unsupervised representation learning. The method adds a split to the network, resulting in two disjoint sub-networks. Each sub-network is trained to perform a difficult task -- predicting one subset of the data channels from another. Together, the sub-networks extract features from the entire input signal. By forcing the network to solve cross-channel prediction tasks, we induce a representation within the network which transfers well to other, unseen tasks. This method achieves state-of-the-art performance on several large-scale transfer learning benchmarks.
\vspace{-3mm}

\end{abstract}

\section{Introduction}

\begin{figure}[t]
 \centering
 \includegraphics[width=1.\hsize]{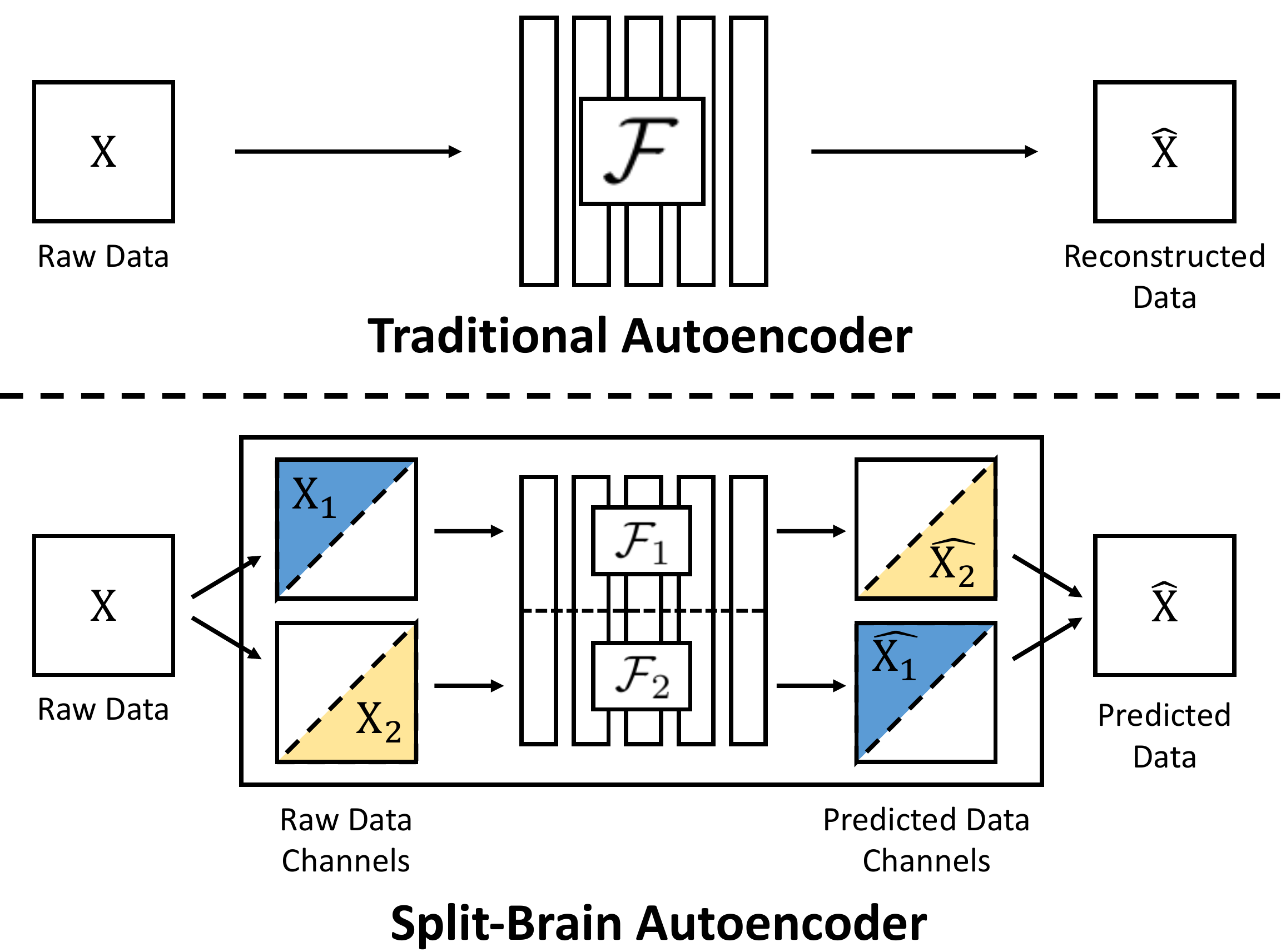}
 \vspace{-4mm}
 \caption{\textbf{Traditional vs Split-Brain Autoencoder architectures.} \textbf{(top)} Autoencoders learn feature representation $\mathcal{F}$ by learning to reconstruct input data $\mathbf{X}$. \textbf{(bottom)} The proposed split-brain autoencoder is composed of two disjoint sub-networks $\mathcal{F}_1,\mathcal{F}_2$, each trained to predict one data subset from another, changing the problem from reconstruction to prediction. The split-brain representation $\mathcal{F}$ is formed by concatenating the two sub-networks, and achieves strong transfer learning performance. The model is publicly available on \url{https://richzhang.github.io/splitbrainauto}.}
 \vspace{-4mm}
\label{fig:fig1}
\end{figure}

A goal of unsupervised learning is to model raw data without the use of labels, in a manner which produces a useful representation. By ``useful" we mean a representation that should be easily adaptable for other tasks, unknown during training time. Unsupervised deep methods typically induce representations by training a network to solve an auxiliary or ``pretext'' task, such as the image {\em reconstruction} objective in a traditional autoencoder model, as shown on Figure \ref{fig:fig1}(top).
We instead force the network to solve complementary {\em prediction} tasks by adding a split in the architecture, shown in Figure \ref{fig:fig1} (bottom), dramatically improving transfer performance.

Despite their popularity, autoencoders have actually not been shown to produce strong representations for transfer tasks in practice \cite{vincent2008extracting,pathakCVPR16context}. Why is this?  One reason might be the mechanism for forcing model abstraction. 
To prevent a trivial identity mapping from being learned, a bottleneck is typically built into the autoencoder representation. However, an inherent tension is at play: the smaller the bottleneck, the greater the forced abstraction, but the smaller the information content that can be expressed. 

Instead of forcing abstraction through compression, via a bottleneck in the network architecture, recent work has explored withholding parts of the input during training \cite{vincent2008extracting,pathakCVPR16context,zhang2016colorful}. For example, Vincent et al. \cite{vincent2008extracting} propose denoising autoencoders, trained to remove \textit{iid} noise added to the input. Pathak et al. \cite{pathakCVPR16context} propose \textit{context encoders}, which learn features by training to inpaint large, random contiguous blocks of pixels.
Rather than dropping data in the spatial direction, several works have dropped data in the channel direction, e.g. predicting color channels from grayscale (the colorization task)~\cite{larsson2016learning,zhang2016colorful}.


Context encoders, while an improvement over autoencoders, demonstrate lower performance than competitors on large-scale semantic representation learning benchmarks~\cite{zhang2016colorful}. This may be due to several reasons. First, image synthesis tasks are known to be notoriously difficult to evaluate ~\cite{ramanarayanan2007visual} and the loss function used in \cite{pathakCVPR16context} may not properly capture inpainting quality. Second, the model is trained on images with missing chunks, but applied, at test time, to full images. This causes a ``domain gap" between training and deployment. Third, it could simply be that the inpainting task in \cite{pathakCVPR16context} could be adequately solved without high-level reasoning, instead mostly just copying low and mid-level structure from the surround.

On the other hand, colorization turns out to be a surprisingly effective pretext task for inducing strong feature representations \cite{zhang2016colorful,larsson2017colorization}. 
Though colorization, like inpainting, is a synthesis task, the spatial correspondence between the input and output pairs may enable basic off-the-shelf loss functions to be effective. In addition, the \textit{systematic}, rather than \textit{stochastic} nature of the input corruption removes the pre-training and testing domain gap. Finally, while inpainting may admit reasoning mainly about textural structure, predicting accurate color, e.g., knowing to paint a schoolbus yellow, may more strictly require object-level reasoning and therefore induce stronger semantic representations. Colorization is an example of what we refer to as a \textit{cross-channel encoding} objective, a task which directly predicts one subset of data channels from another.

In this work, we further explore the space of cross-channel encoders by systematically evaluating various channel translation problems and training objectives. Cross-channel encoders, however, face an inherent handicap: different channels of the input data are not treated equally, as part of the data is used for feature extraction and another as the prediction target. In the case of colorization, the network can only extract features from the grayscale image and is blind to color, leaving the color information unused. A qualitative comparison of the different methods, along with their inherent strengths and weaknesses, is summarized in Table \ref{tab:qual_comp}.

\begin{table}[t!]
\centering
\scalebox{0.83} {
\begin{tabular}{lc@{\hskip 2mm}c@{\hskip 2mm}c}
\specialrule{.1em}{.1em}{.1em}
\textbf{} & \textbf{auxiliary} & \textbf{domain} & \textbf{input} \\
 & \textbf{task type} & \textbf{gap} & \textbf{handicap} \\ \hline
Autoencoder \cite{hinton2006reducing} & reconstruction & no & no \\
Denoising autoencoder \cite{vincent2008extracting} & reconstruction & suffers & no \\
Context Encoder \cite{pathakCVPR16context} & prediction & no & suffers \\
Cross-Channel Encoder \cite{zhang2016colorful,larsson2017colorization} & prediction & no & suffers \\
{\bf Split-Brain Autoencoder} & \textbf{prediction} & \textbf{no} & \textbf{no} \\
\specialrule{.1em}{.1em}{.1em}
\end{tabular}
}
\caption{ \textbf{Qualitative Comparison} We summarize various qualitative aspects inherent in several representation learning techniques. \textbf{Auxiliary task type:} pretext task predicated on reconstruction or prediction. \textbf{Domain gap:} gap between the input data during unsupervised pre-training and testing time. \textbf{Input handicap:} input data is systematically dropped out during test time.}
\label{tab:qual_comp}
\vspace{-4mm}
\end{table}

Might there be a way to take advantage of the underlying principle of cross-channel encoders, while being able to extract features from the entire input signal? We propose an architectural modification to the autoencoder paradigm: adding a single split in the network, resulting in two disjoint, concatenated, sub-networks. Each sub-network is trained as a cross-channel encoder, predicting one subset of channels of the input from the other. A variety of auxiliary cross-channel prediction tasks may be used, such as colorization and depth prediction. For example, on RGB images, one sub-network can solve the problem of colorization (predicting $a$ and $b$ channels from the $L$ channel in $Lab$ colorspace), and the other can perform the opposite (synthesizing $L$ from $a,b$ channels). In the RGB-D domain, one sub-network may predict depth from images, while the other predicts images from depth. The architectural change induces the same forced abstraction as observed in cross-channel encoders, but is able to extract features from the full input tensor, leaving nothing on the table.

Our contributions are as follows: \vspace{-2mm}
\begin{itemize}
\itemsep-.2em
\item We propose the split-brain autoencoder, which is composed of concatenated cross-channel encoders, trained using \textit{raw data} as its own supervisory signal.
\item We demonstrate state-of-the-art performance on several semantic representation learning benchmarks in the RGB and RGB-D domains.
\item To gain a better understanding, we perform extensive ablation studies by (i) investigating cross-channel prediction problems and loss functions and (ii) researching alternative aggregation methods for combining cross-channel encoders.
\end{itemize}

\section{Related Work}

Many unsupervised learning methods have focused on modeling raw data using a reconstruction objective. Autoencoders \cite{hinton2006reducing} train a network to reconstruct an input image, using a representation bottleneck to force abstraction. Denoising autoencoders \cite{vincent2008extracting} train a network to undo a random \textit{iid} corruption. Techniques for modeling the probability distribution of images in deep frameworks have also been explored. For example, variational autoencoders (VAEs) \cite{kingma2013auto} employ a variational Bayesian approach to modeling the data distribution. Other probabilistic models include restricted Boltzmann machines (RBMs) \cite{smolensky1986information}, deep Boltzmann machines (DBMs) \cite{salakhutdinov2009deep}, generative adversarial networks (GANs) \cite{goodfellow2014generative}, autoregressive models (Pixel-RNN \cite{van2016pixel} and Pixel-CNN \cite{oord2016conditional}), bidirectional GANs (BiGANs) \cite{donahue2016adversarial} and Adversarially Learned Inference (ALI) \cite{dumoulin2016adversarially}, and real NVP \cite{dinh2016density}. Many of these methods \cite{hinton2006reducing,vincent2008extracting,donahue2016adversarial,dumoulin2016adversarially,salakhutdinov2009deep} have been evaluated for representation learning.

Another form of unsupervised learning, sometimes referred to as ``self-supervised" learning \cite{de1994learning}, has recently grown in popularity. 
Rather than predicting labels annotated by humans, these methods predict pseudo-labels computed from the raw data itself. For example, image colorization \cite{zhang2016colorful,larsson2016learning} has been shown to be an effective pretext task. Other methods generate pseudo-labels from egomotion~\cite{agrawal2015learning,jayaraman2015learning}, video \cite{wang2015unsupervised,misra2016shuffle}, inpainting~\cite{pathakCVPR16context}, co-occurence~\cite{isola2015learning}, context~\cite{doersch2015unsupervised,noroozi2016unsupervised}, and sound~\cite{owens2016ambient,de2003sensory,de1994learning}. Concurrently, Pathak et al.~\cite{pathak2017learning} use motion masks extracted from video data. Also in these proceedings, Larsson et al.~\cite{larsson2017colorization} provide an in-depth analysis of colorization for self-supervision. These methods generally focus on a single supervisory signal and involve some engineering effort. In this work, we show that simply predicting raw data channels with standard loss functions is surprisingly effective, often outperforming previously proposed methods.

The idea of learning representations from multisensory signals also shows up in structure learning~\cite{ando2005framework}, co-training~\cite{blum1998combining}, and multi-view learning~\cite{xu2013survey}. Our method is especially related to \cite{de1994learning,de2003sensory,sohn2014improved}, which use bidirectional data prediction to learn representations from two sensory modalities.

A large body of additional work in computer vision and graphics focuses on image channel prediction as an end in itself, such as
colorization \cite{zhang2016colorful,larsson2016learning,iizuka2016let}, depth prediction \cite{eigen2015predicting}, and surface normal prediction \cite{eigen2015predicting, wang2015designing}.
In contrast, rather than focusing on the graphics problem, we explore its utility for representation learning.

\section{Methods}

\begin{figure*}[t!]
    \centering
    \begin{subfigure}[t]{0.48\textwidth}
        \centering
        \includegraphics[scale=0.26]{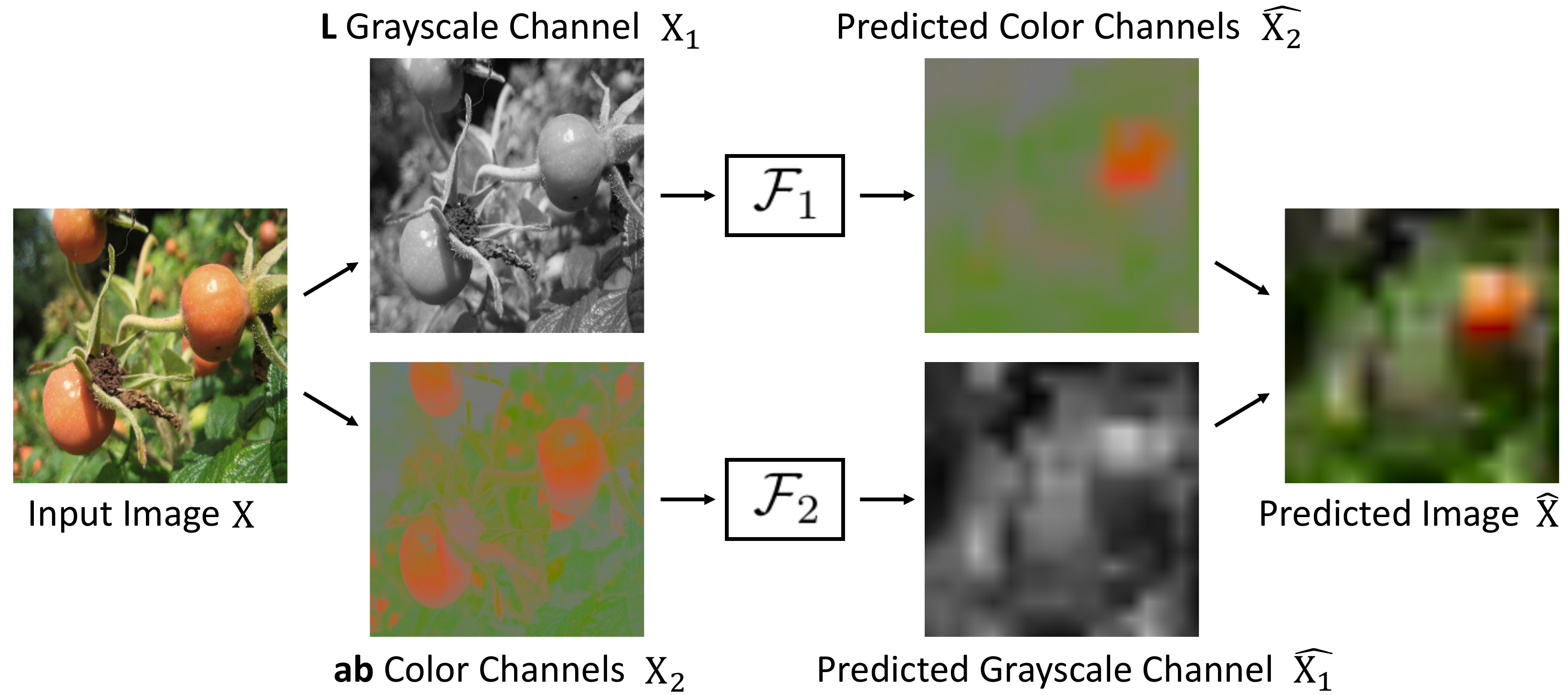}
        \caption{\textbf{\textit{Lab} Images} }
    \end{subfigure}
    \hspace{0.03\textwidth}
    \begin{subfigure}[t]{0.48\textwidth}
        \centering
        \includegraphics[scale=0.26]{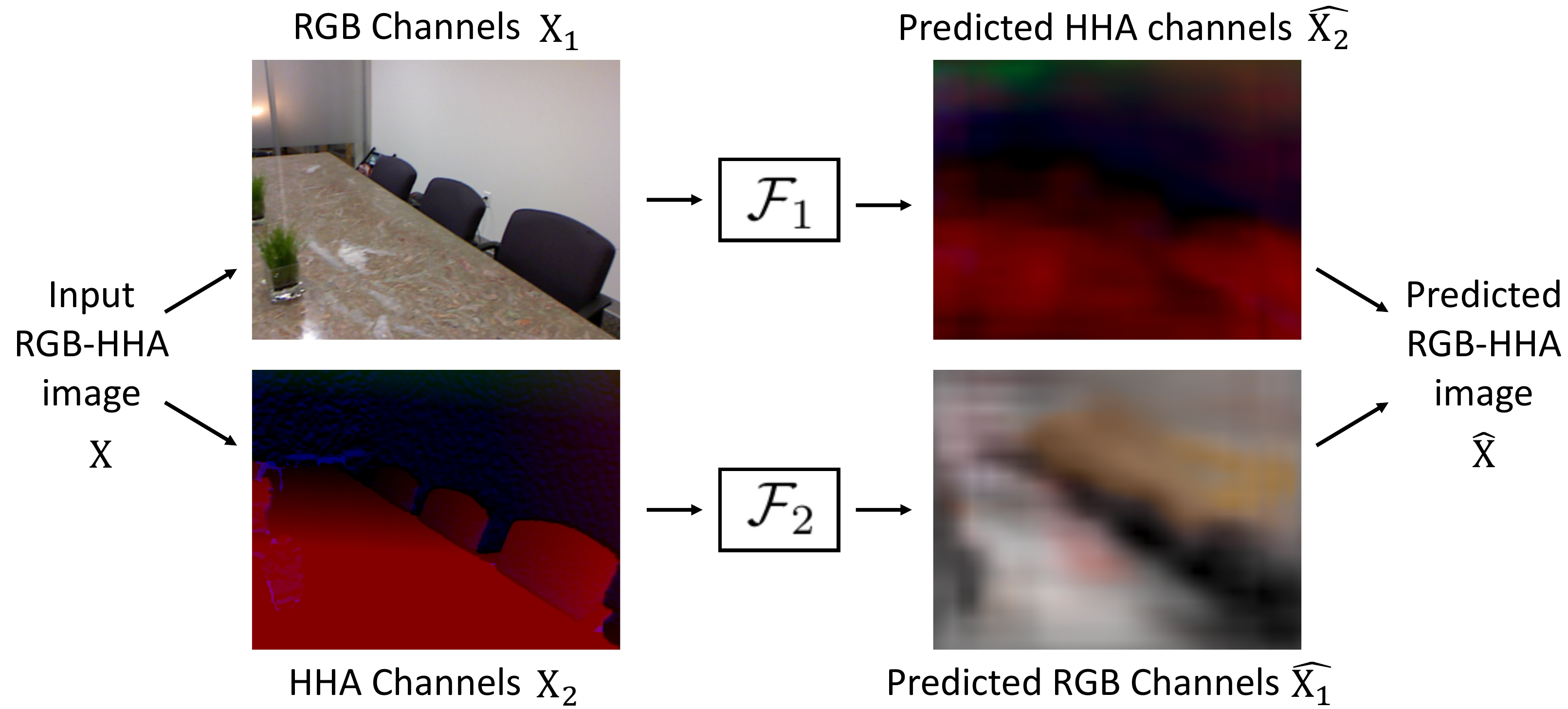}
        \caption{\textbf{RGB-D Images} }
    \end{subfigure}
    \vspace{-1mm}
    \caption{\textbf{Split-Brain Autoencoders applied to various domains} (a) \textbf{\textit{Lab} images} Input images are divided into the $L$ channel, which contains grayscale information, and the $a$ and $b$ channels, which contain color information. Network $\mathcal{F}_1$ performs automatic colorization, whereas network $\mathcal{F}_2$ performs grayscale prediction. (b) \textbf{RGB-D images} Input data $\mathbf{X}$ contains registered RGB and depth images. Depth images are encoded using the HHA encoding \cite{gupta2015cross}. Image representation $\mathcal{F}_1$ is trained by predicting HHA channels. Representation $\mathcal{F}_2$ on HHA images is learned by predicting images in $Lab$ space. Note that the goal of performing these synthesis tasks is to induce representations $\mathcal{F}_1,\mathcal{F}_2$ that transfer well to other tasks.}
    \label{fig:fig2}
\vspace{-3mm}
\end{figure*}

In Section \ref{sec:cross-channel}, we define the paradigm of cross-channel encoding. In Section \ref{sec:agg}, we propose the split-brain autoencoder and explore alternatives methods for aggregating multiple cross-channel encoders into a single network.

\subsection{Cross-Channel Encoders}
\label{sec:cross-channel}

We would like to learn a deep representation on input data tensor $\mathbf{X} \in \mathbb{R}^{H\times W \times C}$, with $\mathbf{C}$ channels. We split the data into $\mathbf{X_1}\in\mathbb{R}^{H\times W \times C_1}$ and $\mathbf{X_2}\in\mathbb{R}^{H\times W\times C_2}$, where $C_1,C_2\subseteq C$, and then train a deep representation to solve the prediction problem $\widehat{\mathbf{X_2}} = \mathcal{F}(\mathbf{X_1})$. Function $\mathcal{F}$ is learned with a CNN, which produces a layered representation of input $\mathbf{X_1}$, and we refer to each layer $l$ as $\mathcal{F}^l$. By performing this \textit{pretext} task of predicting $\mathbf{X_2}$ from $\mathbf{X_1}$, we hope to achieve a representation $\mathcal{F}(\mathbf{X}_1)$ which contains high-level abstractions or semantics.

This prediction task can be trained with various loss functions, and
we study whether the loss function affects the quality of the learned representation. To begin, we explore the use of $\ell_2$ regression, as shown in Equation \ref{eqn:l2loss}.


\vspace{-3mm}
\begin{equation}
\ell_{2}(\mathcal{F}(\mathbf{X_1}),\mathbf{X_2}) = \tfrac{1}{2} \sum_{h,w} \norm{\mathbf{X_2}_{h,w}-\mathcal{F}(\mathbf{X_1})_{h,w}}_2^2
\label{eqn:l2loss}
\end{equation}
\vspace{-3mm}

We also study the use of a classification loss. Here, the target output $\mathbf{X_2}\in\mathbb{R}^{H\times W\times C_2}$ is encoded with function $\mathcal{H}$ into a \textit{distribution} $\mathbf{Y_2}\in \mathbf{\Delta}^{H\times W\times Q}$, where $Q$ is the number of elements in the quantized output space. Network $\mathcal{F}$ is then trained to predict a distribution, $\mathbf{\widehat{Y_2}} = \mathcal{F}(\mathbf{X_1})\in \mathbf{\Delta}^{H\times W\times Q}$. A standard cross-entropy loss between the predicted and ground truth distributions is used, as shown Equation \ref{eqn:classloss}.

\vspace{-5mm}
\begin{equation}
\ell_{cl}(\mathcal{F}(\mathbf{X_1}),\mathbf{X_2}) = - \sum_{h,w}\sum_{q} \mathcal{H}(\mathbf{X_2})_{h,w,q} \log (\mathcal{F}(\mathbf{X_1})_{h,w,q})
\label{eqn:classloss}
\end{equation}

In \cite{zhang2016colorful}, the authors discover that classification loss is more effective for the graphics task of automatic colorization than regression. We hypothesize that for some tasks, especially those with inherent uncertainty in the prediction, the classification loss may lead to better representations as well, as the network will be incentivized to match the whole distribution, and not only predict the first moment.

Note that with input and output sets $C_1,C_2=C$, and an $\ell_2$ regression loss, the objective becomes identical to the autoencoder objective.

\subsection{Split-Brain Autoencoders as Aggregated Cross-Channel Encoders}
\label{sec:agg}

We can train multiple cross-channel encoders, $\mathcal{F}_1$, $\mathcal{F}_2$, on opposite prediction problems, with loss functions $L_1, L_2$, respectively, described in Equation \ref{eqn:concat}.

\vspace{-4mm}
\begin{equation}
\begin{split}
    \mathcal{F}_1^* = \arg\min_{\mathcal{F}_1} L_1(\mathcal{F}_1(\mathbf{X_1}),\mathbf{X_2}) \\
    \mathcal{F}_2^* = \arg\min_{\mathcal{F}_2} L_2(\mathcal{F}_2(\mathbf{X_2}),\mathbf{X_1})
\end{split}
\label{eqn:concat}
\end{equation}
\vspace{-4mm}

By concatenating the representations layer-wise, $\mathcal{F}^l=\{\mathcal{F}^l_1,\mathcal{F}^l_2\}$, we achieve a representation $\mathcal{F}$ which is pre-trained on full input tensor $\mathbf{X}$. Example split-brain autoencoders in the image and RGB-D domains are shown in Figures \ref{fig:fig2}(a) and (b), respectively.
If $\mathcal{F}$ is a CNN of a desired fixed size, e.g., AlexNet \cite{krizhevsky2012imagenet}, we can design the sub-networks $\mathcal{F}_1,\mathcal{F}_2$ by splitting each layer of the network $\mathcal{F}$ in half, along the channel dimension. Concatenated representation $\mathcal{F}$ will then have the appropriate dimensionality, and can be simply implemented by setting the \texttt{group} parameter to 2 in most deep learning libraries. As each channel in the representation is only connected to half of the channels in the preceding layer, the number of parameters in the network is actually halved, relative to a full network.

Note that the input and the output to the network $\mathcal{F}$ is the full input $\mathbf{X}$, the same as an autoencoder. However, due to the split nature of the architecture, the network $\mathcal{F}$ is trained to \textit{predict} $\mathbf{X}=\{\mathbf{X_1},\mathbf{X_2}\}$, rather than simply \textit{reconstruct} it from the input. In essence, an architectural change in the autoencoder framework induces the same forced abstraction achieved by cross-channel encoding.

\textbf{Alternative Aggregation Technique} We found the split-brain autoencoder, which aggregates cross-channel encoders through concatenation, to be more effective than several alternative strategies. As a baseline, we also explore an alternative: the same representation $\mathcal{F}$ can be trained to perform both mappings simultaneously. The loss function is described in Equation \ref{eqn:stoch_comb}, with a slight abuse of notation: here, we redefine $\mathbf{X_1}$ to be the same shape as original input $\mathbf{X}\in\mathbb{R}^{H\times W\times C}$, with channels in set $C\backslash C_1$ zeroed out (along with the analogous modification to $\mathbf{X_2}$).

\vspace{-5mm}
\begin{equation}
\mathcal{F}^* = \arg\min_{\mathcal{F}}  L_1(\mathcal{F}(\mathbf{X_1}),\mathbf{X_2}) + L_2(\mathbf{X_1},\mathcal{F}(\mathbf{X_2}))
\label{eqn:stoch_comb}
\end{equation}
The network only sees data subsets but never full input $\mathbf{X}$. To alleviate this problem, we mix in the autoencoder objective, as shown in Equation \ref{eqn:stoch_comb_auto}, with $\lambda \in [0,\tfrac{1}{2}]$.

\vspace{-4mm}
\begin{equation}
\begin{split}
\mathcal{F}^* = \arg\min_{\mathcal{F}} & \hspace{1mm}
\lambda L_1(\mathcal{F}(\mathbf{X_1}),\mathbf{X_2}) + \lambda L_2(\mathcal{F}(\mathbf{X_2}),\mathbf{X_1}) \\
& + (1-2\lambda) L_3(\mathbf{X},\mathcal{F}(\mathbf{X}))
\end{split}
\label{eqn:stoch_comb_auto}
\end{equation}
Note that unlike the split-brain architecture, in these objectives, there is a domain gap between the distribution of pre-training data and the full input tensor $\mathbf{X}$.

\section{Experiments}

In Section \ref{sec:exp-imnet}, we apply our proposed split-brain autoencoder architecture to learn unsupervised representations on large-scale image data from ImageNet \cite{russakovsky2015imagenet}. We evaluate on established representation learning benchmarks and demonstrate state-of-the-art performance relative to previous unsupervised methods \cite{krahenbuhl2015data,doersch2015unsupervised,wang2015unsupervised,pathakCVPR16context,owens2016ambient,donahue2016adversarial,misra2016shuffle}. In Section \ref{sec:exp-rgbd}, we apply the proposed method on the NYU-D dataset \cite{silberman2012indoor}, and show performance above baseline methods.

\subsection{Split-Brain Autoencoders on Images}
\label{sec:exp-imnet}

We work with image data $\mathbf{X}$ in the \textit{Lab} color space, and learn cross-channel encoders with $\mathbf{X}_1$ representing the $L$, or lightness channel, and $\mathbf{X}_2$ containing the $ab$ channels, or color information. This is a natural choice as (i) networks such as Alexnet, trained with grouping in their architecture, naturally separate into grayscale and color \cite{krizhevsky2012imagenet} even in a fully-supervised setting, and (ii) the individual cross-channel prediction problem of colorization, $L$ to $ab$, has produced strong representations \cite{zhang2016colorful,larsson2016learning}. In preliminary experiments, we have also explored different cross-channel prediction problems in other color spaces, such as \textit{RGB} and \textit{YUV}. We found the \textit{L} and \textit{ab} to be most effective data split. 

To enable comparisons to previous unsupervised techniques, all of our trained networks use AlexNet architectures \cite{krizhevsky2012imagenet}. Concurrent work from Larsson et al.~\cite{larsson2017colorization} shows large performance improvements for the colorization task when using deeper networks, such as VGG-16~\cite{simonyan2014very} and ResNet~\cite{he2015deep}. Because we are training for a pixel-prediction task, we run the network fully convolutionally~\cite{long2015fully}. Using the 1.3M ImageNet dataset \cite{russakovsky2015imagenet} (without labels), we train the following aggregated cross-channel encoders:

\vspace{-2mm}
\begin{itemize}
\itemsep-.1em
\item \textbf{Split-Brain Autoencoder (cl,cl) (Our full method)}: 
A split-brain autoencoder, with one half performing colorization, and the other half performing grayscale prediction. The top-level architecture is shown in Figure \ref{fig:fig2}(a). Both sub-networks are trained for classification (cl), with a cross-entropy objective. (In Figure \ref{fig:fig2}(a), the predicted output is a per-pixel probability distribution, but is visualized with a point estimate using the annealed-mean \cite{zhang2016colorful}.)
\item \textbf{Split-Brain Autoencoder (reg,reg)}: Same as above, with both sub-networks trained with an $\ell_2$ loss (reg).
\item \textbf{Ensembled L$\rightarrow$ab}: Two concatenated disjoint sub-networks, both performing colorization (predicting $ab$ from $L$). One subnetwork is trained with a classification objective, and the other with regression.
\item \textbf{(L,ab)}$\rightarrow$\textbf{(ab,L)}: A single network for both colorization and grayscale prediction, with regression loss, as described in Equation \ref{eqn:stoch_comb}. This explores an alternative method for combining cross-channel encoders.
\item \textbf{(L,ab,Lab)}$\rightarrow$\textbf{(ab,L,Lab)}: $\lambda = \tfrac{1}{3}$ using Equation \ref{eqn:stoch_comb_auto}.
\end{itemize}

Single cross-channel encoders are ablations of our main method. We systematically study combinations of loss functions and cross-channel prediction problems.
\vspace{-1mm}
\begin{itemize}
\itemsep-.2em
\item \textbf{L$\rightarrow$ab(reg)}: Automatic colorization using an $\ell_2$ loss.
\item \textbf{L$\rightarrow$ab(cl)}: Automatic colorization using a classification loss. We follow the quantization procedure proposed in \cite{zhang2016colorful}: the output $ab$ space is binned into grid size $10\times 10$, with a classification loss over the $313$ bins that are within the $ab$ gamut.
\item \textbf{ab$\rightarrow$L(reg)}: Grayscale prediction using an $\ell_2$ loss.
\item \textbf{ab$\rightarrow$L(cl)}: Grayscale prediction using a classification loss. The $L$ channel, which has values between 0 and 100, is quantized into 50 bins of size 2 and encoded.
\item \textbf{Lab$\rightarrow$Lab}: Autoencoder objective, reconstructing $Lab$ from itself using an $\ell_2$ regression loss, with the same architecture as the cross-channel encoders.
\item \textbf{Lab(drop50)$\rightarrow$Lab}: Same as above, with 50\% of the input randomly dropped out during pre-training. This is similar to denoising autoencoders \cite{vincent2008extracting}.
\end{itemize}
\vspace{-1mm}

\begin{table}[t!]
\centering
\scalebox{0.8} {
\begin{tabular}{lccccc}
\specialrule{.1em}{.1em}{.1em}
\multicolumn{6}{c}{\textbf{Task Generalization on ImageNet Classification \cite{russakovsky2015imagenet}}} \\ \specialrule{.1em}{.1em}{.1em}
\specialrule{.1em}{.1em}{.1em}
\textbf{Method} & \textbf{conv1} & \textbf{conv2} & \textbf{conv3} & \textbf{conv4} & \textbf{conv5} \\ \hline 
ImageNet-labels \cite{krizhevsky2012imagenet} & 19.3 & 36.3 & 44.2 & 48.3 & 50.5  \\ \hline 
Gaussian & 11.6 & 17.1 & 16.9 & 16.3 & 14.1  \\
Kr\"ahenb\"uhl et al. \cite{krahenbuhl2015data} & 17.5 & 23.0 & 24.5 & 23.2 & 20.6  \\ \hline
$^{1}$Noroozi \& Favaro \cite{noroozi2016unsupervised} & 19.2 & 30.1 & 34.7 & 33.9 & 28.3  \\ \hline 
Doersch et al. \cite{doersch2015unsupervised} & 16.2 & 23.3 & 30.2 & 31.7 & 29.6  \\
Donahue et al. \cite{donahue2016adversarial} & \textbf{17.7} & 24.5 & 31.0 & 29.9 & 28.0  \\
Pathak et al. \cite{pathakCVPR16context} & 14.1 & 20.7 & 21.0 & 19.8 & 15.5  \\
Zhang et al. \cite{zhang2016colorful} & 13.1 & 24.8 & 31.0 & 32.6 & 31.8  \\ \hline Lab$\rightarrow$Lab & 12.9 & 20.1 & 18.5 & 15.1 & 11.5  \\
Lab(drop50)$\rightarrow$Lab & 12.1 & 20.4 & 19.7 & 16.1 & 12.3  \\
L$\rightarrow$ab(cl) & 12.5 & 25.4 & 32.4 & 33.1 & 32.0  \\
L$\rightarrow$ab(reg) & 12.3 & 23.5 & 29.6 & 31.1 & 30.1  \\
ab$\rightarrow$L(cl) & 11.6 & 19.2 & 22.6 & 21.7 & 19.2  \\
ab$\rightarrow$L(reg) & 11.5 & 19.4 & 23.5 & 23.9 & 21.7  \\ \hline 
(L,ab)$\rightarrow$(ab,L) & 15.1 & 22.6 & 24.4 & 23.2 & 21.1  \\
(L,ab,Lab)$\rightarrow$(ab,L,Lab) & 15.4 & 22.9 & 24.0 & 22.0 & 18.9  \\
Ensembled L$\rightarrow$ab & 11.7 & 23.7 & 30.9 & 32.2 & 31.3  \\
Split-Brain Auto (reg,reg) & 17.4 & 27.9 & 33.6 & 34.2 & 32.3  \\
Split-Brain Auto (cl,cl) & \textbf{17.7} & \textbf{29.3} & \textbf{35.4} & \textbf{35.2} & \textbf{32.8}  \\
\specialrule{.1em}{.1em}{.1em}
\end{tabular}
\vspace{-2mm}
}
\caption {\textbf{Task Generalization on ImageNet Classification} To test unsupervised feature representations, we train linear logistic regression classifiers on top of each layer to perform 1000-way ImageNet classification, as proposed in \cite{zhang2016colorful}. All weights are frozen and feature maps spatially resized to be $\sim$9000 dimensions. All methods use AlexNet variants \cite{krizhevsky2012imagenet}, and were pre-trained on ImageNet without labels, except for \textbf{ImageNet-labels}. Note that the proposed split-brain autoencoder achieves the best performance on all layers across unsupervised methods.}
\vspace{-4mm}
\label{tab:ilsvrclin}
\end{table}

\begin{table}[!]
\centering
\scalebox{0.8} {
\begin{tabular}{lccccc}
\specialrule{.1em}{.1em}{.1em}
\multicolumn{6}{c}{\textbf{Dataset \& Task Generalization on Places Classification \cite{zhou2014learning}}} \\ \specialrule{.1em}{.1em}{.1em}
\specialrule{.1em}{.1em}{.1em}
\textbf{Method} & \textbf{conv1} & \textbf{conv2} & \textbf{conv3} & \textbf{conv4} & \textbf{conv5} \\ \hline 
Places-labels \cite{zhou2014learning} & 22.1 & 35.1 & 40.2 & 43.3 & 44.6  \\
ImageNet-labels \cite{krizhevsky2012imagenet} & 22.7 & 34.8 & 38.4 & 39.4 & 38.7  \\ \hline 
Gaussian & 15.7 & 20.3 & 19.8 & 19.1 & 17.5  \\
Kr\"ahenb\"uhl et al. \cite{krahenbuhl2015data} & 21.4 & 26.2 & 27.1 & 26.1 & 24.0  \\ \hline
$^{1}$Noroozi \& Favaro \cite{noroozi2016unsupervised} & 23.0 & 32.1 & 35.5 & 34.8 & 31.3  \\ \hline 
Doersch et al. \cite{doersch2015unsupervised} & 19.7 & 26.7 & 31.9 & 32.7 & 30.9  \\
Wang \& Gupta \cite{wang2015unsupervised} & 20.1 & 28.5 & 29.9 & 29.7 & 27.9  \\
Owens et al. \cite{owens2016ambient} & 19.9 & 29.3 & 32.1 & 28.8 & 29.8  \\
Donahue et al. \cite{donahue2016adversarial} & \textbf{22.0} & 28.7 & 31.8 & 31.3 & 29.7  \\
Pathak et al. \cite{pathakCVPR16context} & 18.2 & 23.2 & 23.4 & 21.9 & 18.4  \\
Zhang et al. \cite{zhang2016colorful} & 16.0 & 25.7 & 29.6 & 30.3 & 29.7  \\ \hline
L$\rightarrow$ab(cl) & 16.4 & 27.5 & 31.4 & 32.1 & 30.2  \\
L$\rightarrow$ab(reg) & 16.2 & 26.5 & 30.0 & 30.5 & 29.4  \\
ab$\rightarrow$L(cl) & 15.6 & 22.5 & 24.8 & 25.1 & 23.0  \\
ab$\rightarrow$L(reg) & 15.9 & 22.8 & 25.6 & 26.2 & 24.9  \\ \hline
Split-Brain Auto (cl,cl) & 21.3 & \textbf{30.7} & \textbf{34.0} & \textbf{34.1} & \textbf{32.5}  \\
\specialrule{.1em}{.1em}{.1em}
\end{tabular}
}
\caption{\textbf{Dataset \& Task Generalization on Places Classification} 
We train logistic regression classifiers on top of frozen pre-trained representations for 205-way Places classification. Note that our split-brain autoencoder achieves the best performance among unsupervised learning methods from \texttt{conv2-5} layers.}
\vspace{-2mm}
\label{tab:placeslin}
\end{table}

We compare to the following methods, which all use variants of Alexnet \cite{krizhevsky2012imagenet}. For additional details, refer to Table 3 in \cite{zhang2016colorful}. Note that one of these modifications resulted in a large deviation in feature map size\footnote{The method from \cite{noroozi2016unsupervised} uses stride 2 instead of 4 in the \texttt{conv1} layer, resulting in 4$\times$ denser feature maps throughout all convolutional layers. While it is unclear how this change affects representational quality, experiments from Larsson et al. \cite{larsson2017colorization} indicate that changes in architecture can result in large changes in transfer performance, even given the same training task. The network uses the same number of parameters, but 5.6$\times$ the memory and 7.4$\times$ the run-time.}.
\begin{itemize}
\itemsep-.4em
\item \textbf{ImageNet-labels} \cite{krizhevsky2012imagenet}: Trained on ImageNet labels for the classification task in a fully supervised fashion.
\item \textbf{Gaussian}: Random Gaussian initialization of weights.
\item \textbf{Kr\"ahenb\"uhl et al.} \cite{krahenbuhl2015data}: A stacked k-means initialization method.
\item \textbf{Doersch et al.} \cite{doersch2015unsupervised}, \textbf{Noroozi \& Favaro} \cite{noroozi2016unsupervised}, \textbf{Pathak et al.} \cite{pathakCVPR16context}, \textbf{Donahue et al.} \cite{donahue2016adversarial}, and \textbf{Zhang et al.} \cite{zhang2016colorful} all pre-train on the 1.3M ImageNet dataset \cite{russakovsky2015imagenet}.
\item \textbf{Wang \& Gupta} \cite{wang2015unsupervised} and \textbf{Owens et al.} \cite{owens2016ambient} pre-train on other large-scale data.
\end{itemize}
\vspace{-3mm}

\begin{figure*}[t!]
    \hspace{-0.07\textwidth}
    \centering
    \begin{subfigure}[t]{0.47\textwidth}
        \centering
        \includegraphics[scale=0.45]{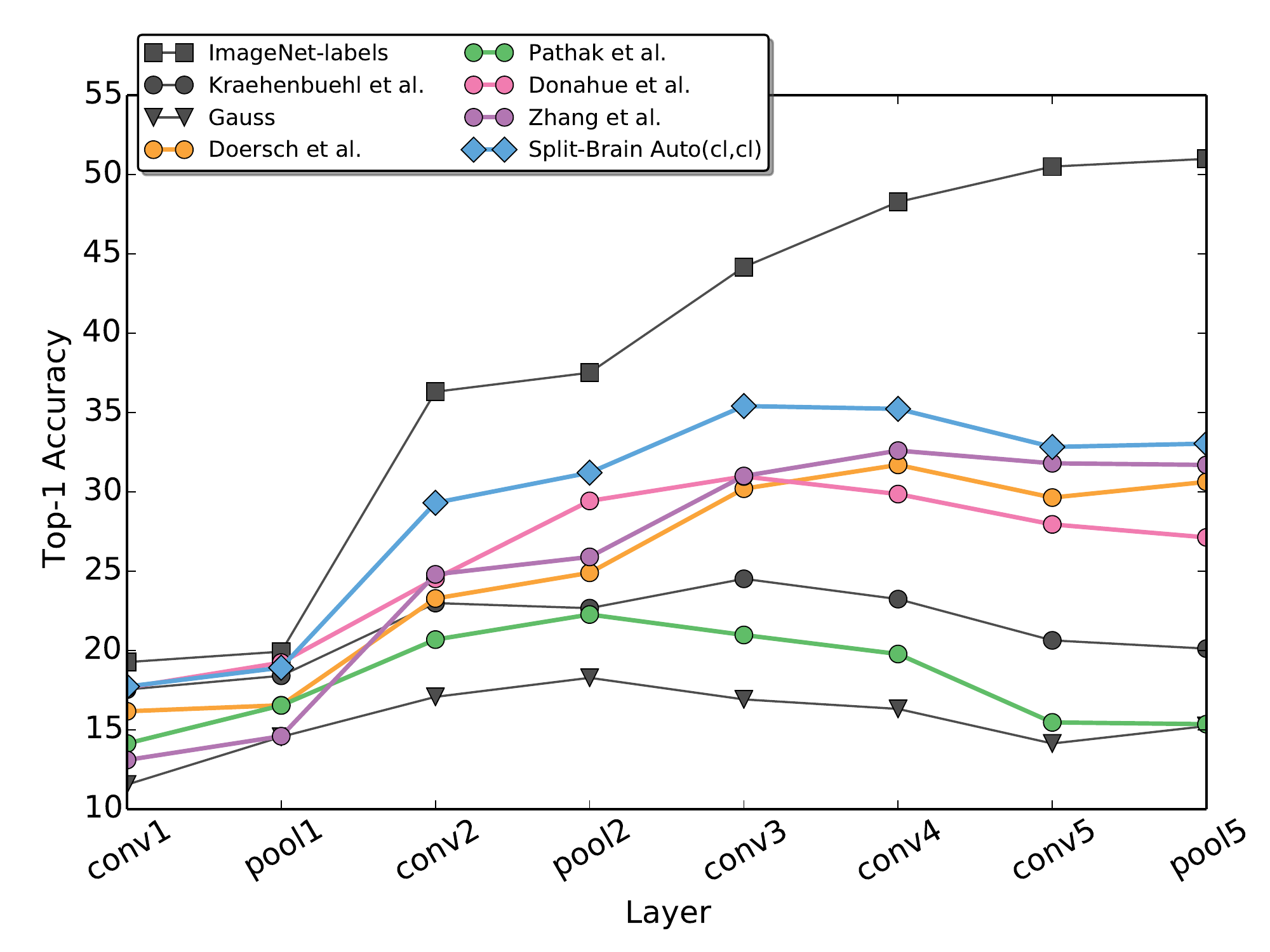}
        \caption{\textbf{ImageNet Classification} }
    \end{subfigure}
    \hspace{0.06\textwidth}
    \begin{subfigure}[t]{0.47\textwidth}
        \centering
        \includegraphics[scale=0.45]{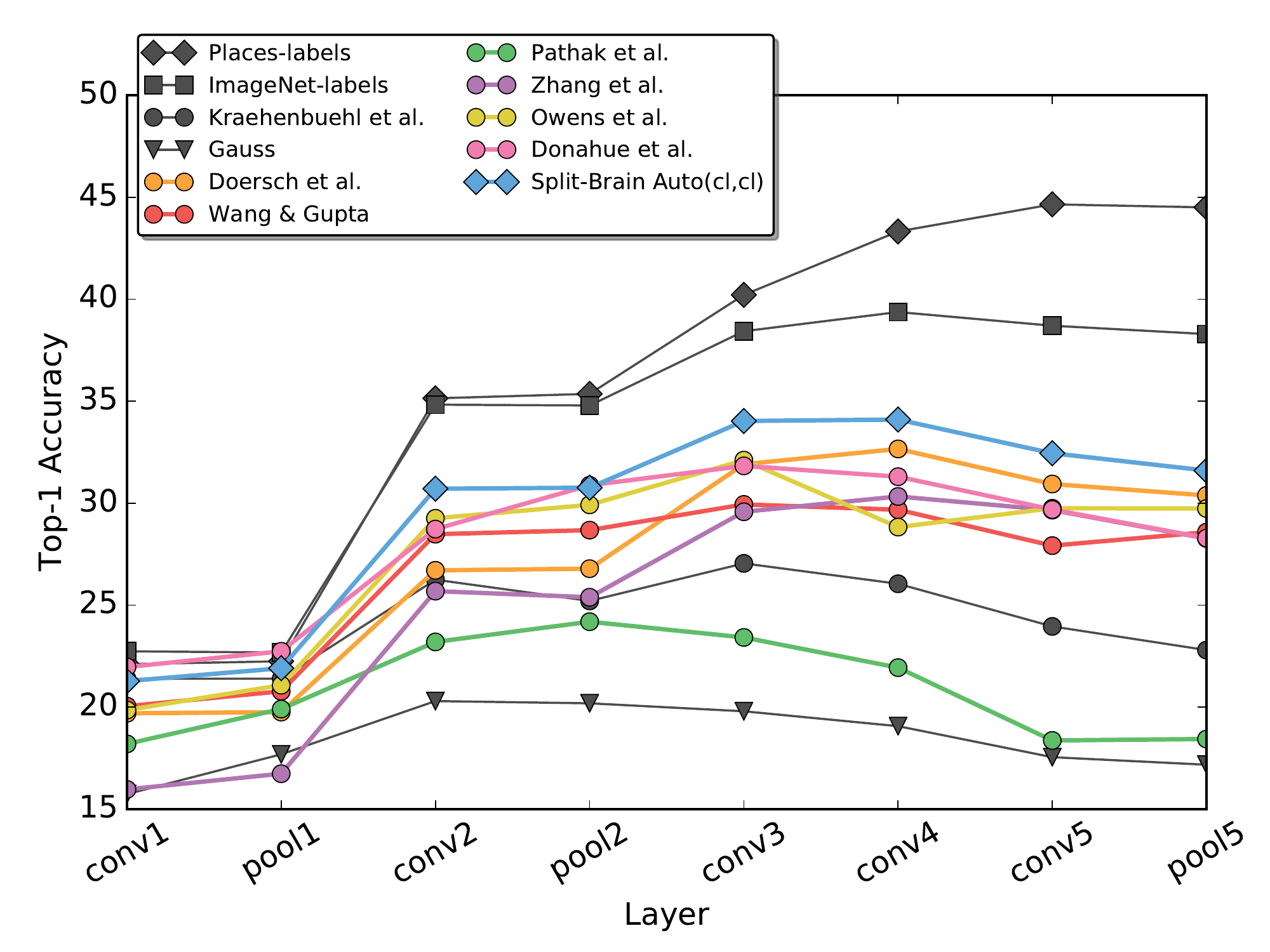}
        \caption{\textbf{Places Classification} }
    \end{subfigure}
    \vspace{-1mm}
    \caption{\textbf{Comparison to Previous Unsupervised Methods} We compare our proposed Split-Brain Autoencoder on the tasks of (a) ImageNet classification and (b) Places Classification. Note that our method outperforms other large-scale unsupervised methods \cite{doersch2015unsupervised,wang2015unsupervised,pathakCVPR16context,zhang2016colorful,owens2016ambient,donahue2016adversarial} on all layers in ImageNet and from \texttt{conv2-5} on Places.}
    \label{fig:lin_method_comp}
\vspace{-3mm}
\end{figure*}

\begin{figure*}[t!]
    \begin{subfigure}[t]{0.33\textwidth}
        \centering
        \includegraphics[scale=0.45]{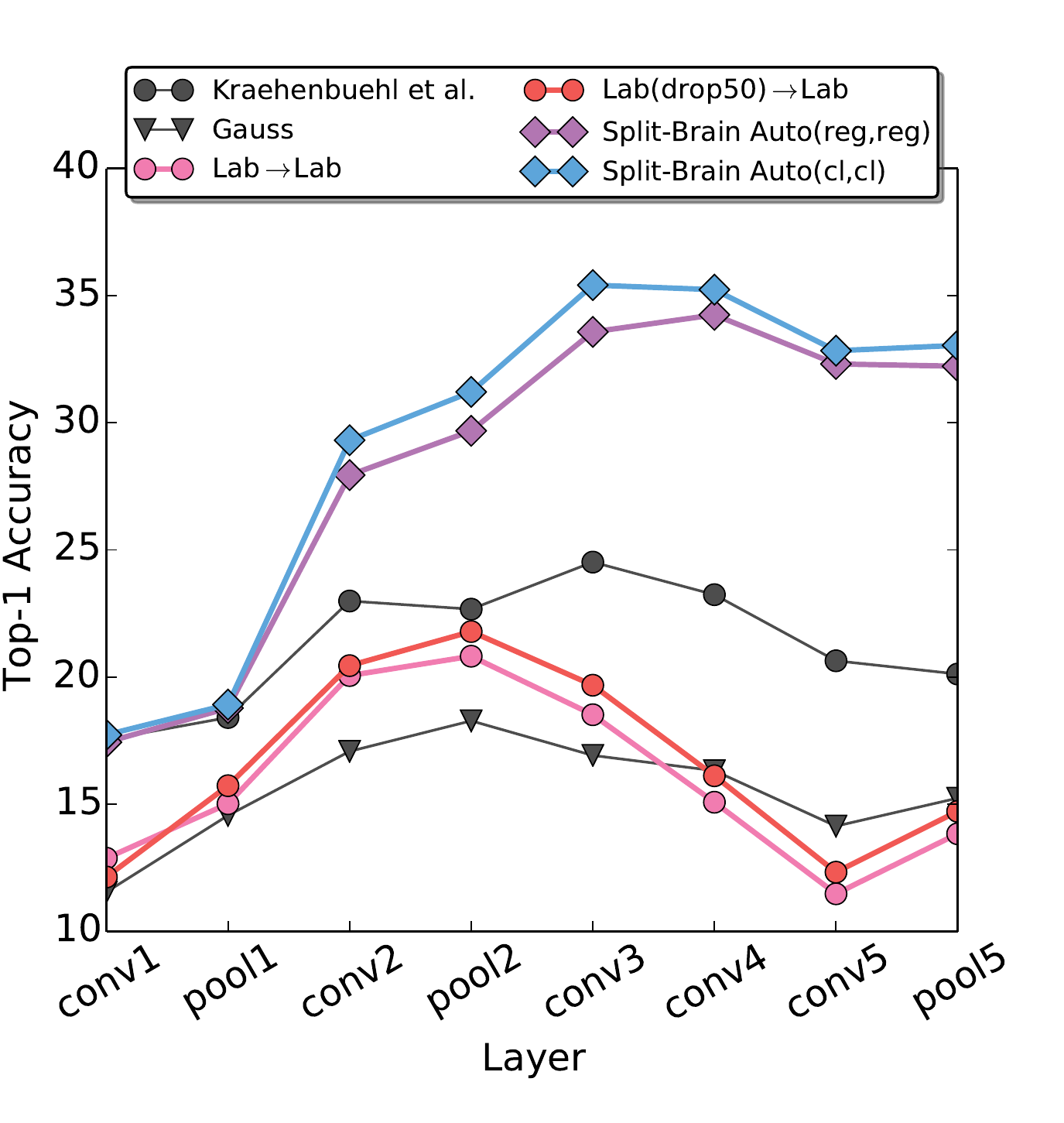}
        \caption{\textbf{Autoencoder Objective} }
    \end{subfigure}
    \hspace{0.01\textwidth}
    \begin{subfigure}[t]{0.33\textwidth}
        \centering
        \includegraphics[scale=0.45]{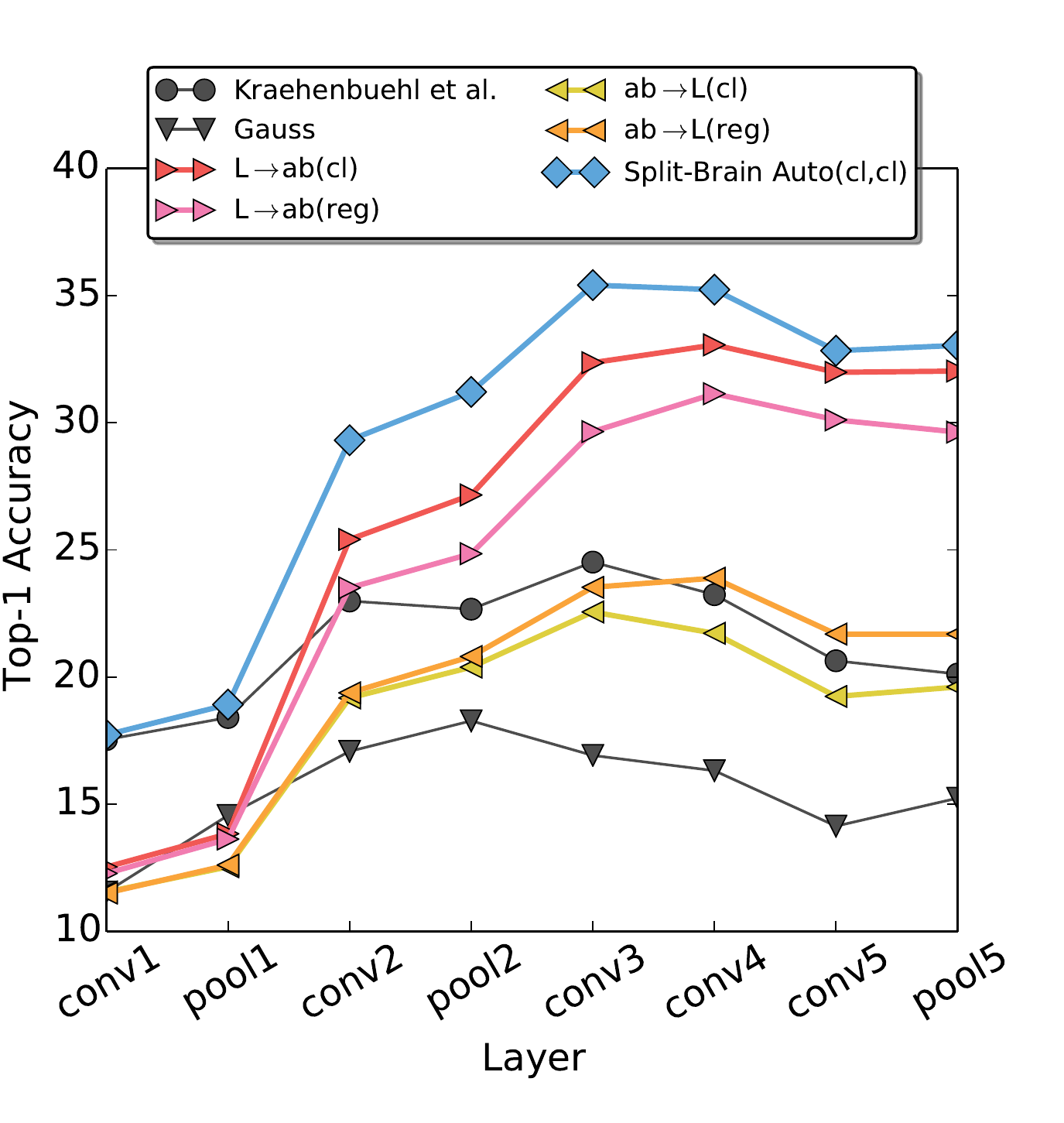}
        \caption{\textbf{Cross-Channel Encoders} }
    \end{subfigure}
    \hspace{0.01\textwidth}
    \begin{subfigure}[t]{0.33\textwidth}
        \centering
        \includegraphics[scale=0.45]{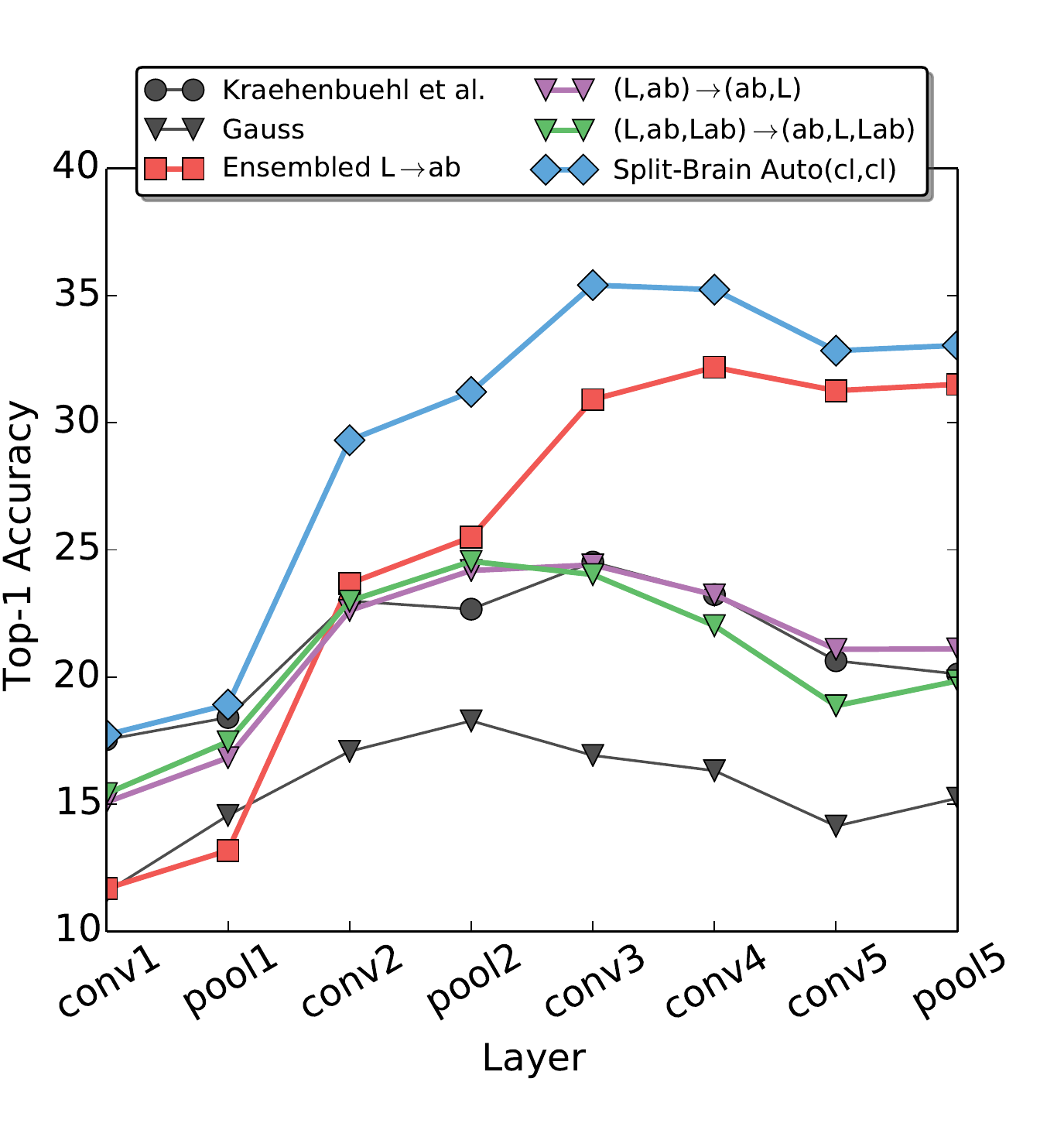}
        \caption{\textbf{Aggregation Methods} }
    \end{subfigure}
    \vspace{-1mm}
    \caption{\textbf{Ablation Studies} We conduct various ablation studies on our proposed method, using the ImageNet classification benchmark proposed in \cite{zhang2016colorful}. Specifically, we compare (a) variations using an autoencoder objective (b) different cross-channel problems and loss functions (c) different methods for aggregating multiple cross-channel encoders.}
    \label{fig:ilsvrclin_comp}
    \vspace{-2mm}
\end{figure*}

\subsubsection{Transfer Learning Tests}

How well does the pre-text task of cross-channel prediction generalize to unseen tasks and data? We run various established large-scale representation learning benchmarks.

\textbf{ImageNet} \cite{krizhevsky2012imagenet} As proposed in \cite{zhang2016colorful}, we test the \textit{task} generalization of the representation by freezing the weights and training multinomial logistic regression classifiers on top of each layer to perform 1000-way ImageNet classification. Note that each classifier is a single learned linear layer, followed by a softmax. To reduce the effect of differences in feature map sizes, we spatially resize feature maps through bilinear interpolation, so that the flattened feature maps have approximately equal dimensionality (9600 for \texttt{conv1,3,4} and 9216 for \texttt{conv2,5}). The results are shown in Table \ref{tab:ilsvrclin} and Figures \ref{fig:lin_method_comp}(a) and \ref{fig:ilsvrclin_comp}.

\textbf{Places} \cite{zhou2014learning} In the previous test, we evaluated the representation on the same input training data, the ImageNet dataset, with a different task than the pretraining tasks. To see how well the network generalizes to new input \textit{data} as well, we run the same linear classification task on the large-scale Places dataset \cite{zhou2014learning}. The dataset contains 2.4M images for training and 20.5k for validation from 205 scene categories. The results are shown in Table \ref{tab:placeslin} and Figure \ref{fig:lin_method_comp}(b).

\textbf{PASCAL} \cite{everingham2010pascal} To further test generalization, we fine-tune the learned representation on standard representation learning benchmarks on the PASCAL dataset, as shown in Table \ref{tab:pascal}, using established testing frameworks in classification \cite{krahenbuhl2015data}, detection \cite{girshick2015fast}, and segmentation \cite{long2015fully}. Classification involves 20 binary classification decisions, regarding the presence or absence of 20 object classes. Detection involves drawing an accurately localized bounding box around any objects in the image, and is performed using the Fast R-CNN \cite{girshick2015fast} framework. Segmentation is pixel-wise labeling of the object class, either one of the 20 objects of interest or background. Here, the representation is \textit{fine-tuned} through multiple layers of the network, rather than frozen. Prior to fine-tuning, we follow common practice and use the rescaling method from \cite{krahenbuhl2015data}, which rescales the weights so that the layers learn at the same ``rate", using the ratio of expected gradient magnitude over feature activation magnitude as a heuristic.

\vspace{-2mm}

\begin{table}[t!]
\hspace{-3mm}
\centering
\scalebox{0.71} {
\begin{tabular}{@{\hskip 1mm}l@{\hskip 4mm}c@{\hskip 2mm}c@{\hskip 2mm}c@{\hskip 9mm}c@{\hskip 2mm}c@{\hskip 9mm}c@{\hskip 2mm}c@{\hskip 1mm}}
\specialrule{.1em}{.1em}{.1em}
\multicolumn{8}{c}{\textbf{Task and Data Generalization on PASCAL VOC \cite{everingham2010pascal}}} \\
\specialrule{.1em}{.1em}{.1em}
\specialrule{.1em}{.1em}{.1em}
 & \multicolumn{3}{c}{\hskip -8mm\textbf{Classification \cite{krahenbuhl2015data}}} & \multicolumn{2}{c}{\hskip -9mm\textbf{Detection \cite{girshick2015fast}}} & \multicolumn{2}{c}{\hskip -4mm\textbf{Seg. \cite{long2015fully}}} \\
 & \multicolumn{3}{c}{\hskip -8mm\textbf{(\%mAP)}} & \multicolumn{2}{c}{\hskip -9mm\textbf{(\%mAP)}} & \multicolumn{2}{c}{\hskip -4mm\textbf{(\%mIU)}} \\
\specialrule{.1em}{.1em}{.1em}
\textbf{frozen layers} & & \textbf{conv5} & \textbf{none} & & \textbf{none} & & \textbf{none} \\
\textbf{fine-tuned layers} & \textbf{Ref} & \textbf{fc6-8} & \textbf{all} & \textbf{Ref} & \textbf{all} & \textbf{Ref} & \textbf{all} \\ \specialrule{.1em}{.1em}{.1em}
ImageNet labels \cite{krizhevsky2012imagenet} & \cite{zhang2016colorful} & 78.9 & 79.9 & \cite{krahenbuhl2015data} & 56.8 & \cite{long2015fully} & 48.0 \\ \hline
Gaussian & \cite{pathakCVPR16context} & -- & 53.3 & \cite{pathakCVPR16context} & 43.4 & \cite{pathakCVPR16context} & 19.8 \\
Autoencoder & \cite{donahue2016adversarial} & 16.0 & 53.8 & \cite{pathakCVPR16context} & 41.9 & \cite{pathakCVPR16context} & 25.2 \\
Kr\"ahenb\"uhl et al. \cite{krahenbuhl2015data} & \cite{donahue2016adversarial} & 39.2 & 56.6 & \cite{krahenbuhl2015data} & 45.6 & \cite{donahue2016adversarial} & 32.6 \\ \hline
Jayaraman \& Grauman \cite{jayaraman2015learning} & -- & -- & -- & \cite{jayaraman2015learning} & 41.7 & -- & -- \\
Agrawal et al. \cite{agrawal2015learning} & \cite{krahenbuhl2015data} & -- & 52.9 & \cite{krahenbuhl2015data} & 41.8 & -- & -- \\
Agrawal et al. \cite{agrawal2015learning}$^{\dagger}$ & \cite{donahue2016adversarial} & 31.0 & 54.2 & \cite{krahenbuhl2015data} & 43.9 & -- & -- \\
Wang \& Gupta \cite{wang2015unsupervised} & \cite{krahenbuhl2015data} & -- & 62.8 & \cite{krahenbuhl2015data} & 47.4 & -- & -- \\
Wang \& Gupta \cite{wang2015unsupervised}$^{\dagger}$ & \cite{krahenbuhl2015data} & -- & 63.1 & \cite{krahenbuhl2015data} & 47.2 & -- & -- \\
Doersch et al. \cite{doersch2015unsupervised} & \cite{krahenbuhl2015data} & -- & 55.3 & \cite{krahenbuhl2015data} & 46.6 & -- & -- \\
Doersch et al. \cite{doersch2015unsupervised}$^{\dagger}$ & \cite{donahue2016adversarial} & 55.1 & 65.3 & \cite{krahenbuhl2015data} & 51.1 & -- & -- \\
Pathak et al. \cite{pathakCVPR16context} & \cite{pathakCVPR16context} & -- & 56.5 & \cite{pathakCVPR16context} & 44.5 & \cite{pathakCVPR16context} & 29.7 \\
Donahue et al. \cite{donahue2016adversarial}$^{\dagger}$ & \cite{donahue2016adversarial} & 52.3 & 60.1 & \cite{donahue2016adversarial} & 46.9 & \cite{donahue2016adversarial} & 35.2 \\ 
Misra et al. \cite{misra2016shuffle} & -- & -- & -- & \cite{misra2016shuffle} & 42.4 & -- & -- \\
Owens et al. \cite{owens2016ambient} & $\triangleright$ & 54.6 & 54.4 & \cite{owens2016ambient} & 44.0 & -- & -- \\
Owens et al. \cite{owens2016ambient}$^{\dagger}$ & $\triangleright$ & 52.3 & 61.3 & -- & -- & -- & -- \\
Zhang et al. \cite{zhang2016colorful}$^{\dagger}$ & \cite{zhang2016colorful} & 61.5 & 65.9 & \cite{zhang2016colorful} & 46.9 & \cite{zhang2016colorful} & 35.6 \\ \hline
Larsson et al. \cite{larsson2017colorization}$\diamond$ & \cite{larsson2017colorization} & -- & 65.9 & -- & -- & \cite{larsson2017colorization} & \textbf{38.4} \\
Pathak et al. \cite{pathak2017learning}$\diamond$ & \cite{pathak2017learning} & -- & 61.0 & \cite{pathak2017learning} & \textbf{52.2} & -- & -- \\ \hline
Split-Brain Auto (cl,cl)$^{\dagger}$ & $\triangleright$ & \textbf{63.0} & \textbf{67.1} & $\triangleright$ & 46.7 & $\triangleright$ & 36.0 \\
\specialrule{.1em}{.1em}{.1em}
\end{tabular}
}
\caption {\textbf{Task and Dataset Generalization on PASCAL VOC} Classification and detection on PASCAL VOC 2007 \cite{pascal-voc-2007} and segmentation on PASCAL VOC 2012 \cite{pascal-voc-2012}, using mean average precision (mAP) and mean intersection over union (mIU) metrics for each task, with publicly available testing frameworks from \cite{krahenbuhl2015data}, \cite{girshick2015fast}, \cite{long2015fully}. Column \textbf{Ref} documents the source for a value obtained from a previous paper. Character $\triangleright$ indicates that value originates from this paper. $^{\dagger}$indicates that network weights have been rescaled with \cite{krahenbuhl2015data} before fine-tuning, as is common practice. Character $\diamond$ indicates concurrent work in these proceedings.
}
\label{tab:pascal}
\vspace{-5mm}
\end{table}

\subsubsection{Split-Brain Autoencoder Performance}

Our primary result is that the proposed method, \textbf{Split-Brain Auto (cl,cl)}, achieves state-of-the-art performance on almost all established self-supervision benchmarks, as seen in the last row on Tables \ref{tab:ilsvrclin}, \ref{tab:placeslin}, \ref{tab:pascal}, over previously proposed self-supervision methods, as well as our ablation baselines. Figures \ref{fig:lin_method_comp}(a) and (b) shows our split brain autoencoder method compared to previous self-supervised methods \cite{doersch2015unsupervised,wang2015unsupervised,pathakCVPR16context,zhang2016colorful,donahue2016adversarial,owens2016ambient} on the ImageNet and Places classification tests, respectively. We especially note the straightforward nature of our proposed method: the network simply predicts raw data channels from other raw data channels, using a classification loss with a basic 1-hot encoding scheme.

As seen in Figure \ref{fig:ilsvrclin_comp}(a) and Table \ref{tab:ilsvrclin}, the autoencoder objective by itself, \textbf{Lab$\rightarrow$Lab}, does not lead to a strong representation. Performance is near Gaussian initialization through the initial layers, and actually falls below in the \texttt{conv5} layer. Dropping 50\% of the data from the input randomly during training, \textbf{Lab(drop50)$\rightarrow$Lab}, in the style of denoising autoencoders, adds a small performance boost of approximately 1\%. A large performance boost is observed by adding a split in the architecture, \textbf{Split-Brain Auto (reg,reg)}, even with the same regression objective. This achieves 5\% to 20\% higher performance throughout the network, state-of-the-art compared to previous unsupervised methods. A further boost of approximately 1-2\% throughout the network observed using a classification loss, \textbf{Split-Brain Auto (cl,cl)}, instead of regression. 
\vspace{-2mm}

\subsubsection{Cross-Channel Encoding Objectives}

Figure \ref{fig:ilsvrclin_comp}(b) compares the performance of the different cross-channel objectives we tested on the ImageNet classification benchmark.
As shown in \cite{zhang2016colorful} and further confirmed here, colorization, \textbf{L$\rightarrow$ab(cl)}, leads to a strong representation on classification transfer tasks, with higher performance than other unsupervised representations pre-trained on ImageNet, using inpainting \cite{pathakCVPR16context}, relative context \cite{doersch2015unsupervised}, and adversarial feature networks \cite{donahue2016adversarial} from layers from \texttt{conv2} to \texttt{pool5}. We found that the classification loss produced stronger representations than regression for colorization, consistent with the findings from concurrent work from Larsson et al. \cite{larsson2017colorization}.

Interestingly, the task of predicting grayscale from color can also learn representations. Though colorization lends itself closely to a graphics problem, the application of grayscale prediction from color channels is less obvious. As seen in Tables \ref{tab:ilsvrclin} and \ref{tab:placeslin} and Figure \ref{fig:ilsvrclin_comp}(b), grayscale prediction objectives \textbf{ab$\rightarrow$L(cl)} and \textbf{ab$\rightarrow$L(reg)} can learn representations above the \textbf{Gaussian} baseline. Though the learned representation by itself is weaker than other self-supervised methods, the representation is learned on $a$ and $b$ channels, which makes it complementary to the colorization network. For grayscale prediction, regression results in higher performance than classification. Choosing the appropriate loss function for a given channel prediction problem is an open problem. However, note that the performance difference is typically small, indicating that the cross-channel prediction problem is often times an effective method, even without careful engineering of the objective.


\subsection{Split-Brain Autoencoders on RGB-D}
\label{sec:exp-rgbd}

\begin{table}
\centering
\scalebox{0.7} {
\begin{tabular}{lccccc}
\specialrule{.1em}{.1em}{.1em}
\textbf{Method} & \textbf{Data} & \textbf{Label} & \textbf{RGB} & \textbf{D} & \textbf{RGB-D} \\ \hline
Gupta et al. \cite{gupta2015cross} & 1M ImNet \cite{russakovsky2015imagenet} & \checkmark & 27.8 & 41.7 & 47.1 \\ Gupta et al. \cite{gupta2014learning} & 1M ImNet \cite{russakovsky2015imagenet} & \checkmark & 27.8 & 34.2 & 44.4 \\ \hline 
Gaussian & None & & -- & 28.1 & -- \\
Kr\"ahenb\"uhl et al. \cite{krahenbuhl2015data} & 20 NYU-D \cite{silberman2012indoor} & & 12.5 & 32.2 & 34.5 \\ \hline
Split-Brain Autoencoder & 10k NYU-D \cite{silberman2012indoor} & & \textbf{18.9} & \textbf{33.2} & \textbf{38.1} \\
\specialrule{.1em}{.1em}{.1em}
\end{tabular} }
\vspace{-1mm}
\caption {\textbf{Split-Brain Autoencoder Results on RGB-D images} We perform unsupervised training on 10k RGB-D keyframes from the NYU-D \cite{silberman2012indoor} dataset, extracted by \cite{gupta2015cross}. We pre-train representations on RGB images using $\ell_2$ loss on depth images in \textit{HHA} space. We pre-train \textit{HHA} representations on $L$ and $ab$ channels using $\ell_2$ and classification loss, respectively. We show performance gains above Gaussian and Kr\"ahenb\"uhl et al. \cite{krahenbuhl2015data} initialization baselines. The methods proposed by Gupta et al. \cite{gupta2014learning,gupta2015cross} use 1.3M labeled images for supervised pre-training. We use the test procedure from \cite{gupta2015cross}: Fast R-CNN \cite{girshick2015fast} networks are first trained individually in the RGB and D domains separately, and then ensembled together by averaging (RGB-D).}
\vspace{-3mm}
\label{tab:rgbd}
\end{table}

We also test the split-brain autoencoder method on registered images and depth scans from NYU-D \cite{silberman2012indoor}. Because RGB and depth images are registered spatially, RGB-D data can be readily applied in our proposed framework. We split the data by modality, predicting RGB from D and vice-versa. Previous work in the video and audio domain \cite{de2003sensory} suggest that separating modalities, rather than mixing them, provides more effective splits. This choice also provides easy comparison to the test procedure introduced by \cite{gupta2014learning}.

\textbf{Dataset \& Detection Testbed} The NYUD dataset contains 1449 RGB-D labeled images and over 400k unlabeled RGB-D video frames. We use 10k of these unlabeled frames to perform unsupervised pre-training, as extracted from \cite{gupta2015cross}. We evaluate the representation on the 1449 labeled images for the detection task, using the framework proposed in \cite{gupta2015cross}. The method first trains individual detectors on the RGB and D domains, using the Fast R-CNN framework \cite{girshick2015fast} on an AlexNet architecture, and then late-fuses them together through ensembling.

\textbf{Unsupervised Pre-training} We represent depth images using the HHA encoding, introduced in \cite{gupta2014learning}. To learn image representation $\mathcal{F}_{HHA}$, we train an Alexnet architecture to regress from RGB channels to HHA channels, using an $\ell_2$ regression loss.


To learn depth representations, we train an Alexnet on HHA encodings, using $\ell_2$ loss on \textit{L} and classification loss on \textit{ab} color channels. We chose this combination, as these objectives performed best for training individual cross-channel encoders in the image domain.
The network extracts features up to the \texttt{conv5} layer, using an Alexnet architecture, and then splits off into specific branches for the $L$ and $ab$ channels. Each branch contains AlexNet-type \texttt{fc6-7} layers, but with 512 channels each, evaluated fully convolutionally for pixel prediction. The loss on the $ab$ term was weighted 200$\times$ with respect to the $L$ term, so the gradient magnitude on the \texttt{pool5} representation from channel-specific branches were approximately equal throughout training.

Across all methods, weights up to the \texttt{conv5} layer are copied over during fine-tuning time, and \texttt{fc6-7} layers are randomly initialized, following \cite{gupta2014learning}.

\textbf{Results}
The results are shown in Table \ref{tab:rgbd} for detectors learned in RGB and D domains separately, as well as the ensembled result. For a Gaussian initialization, the RGB detector did not train using default settings, while the depth detector achieved performance of 28.1\%. Using the stacked k-means initialization scheme from Kr\"ahenb\"uhl et al.~\cite{krahenbuhl2015data}, individual detectors on RGB and D perform at 12.5\% and 32.2\%, while achieving 34.5\% after ensembling. Pre-training with our method reaches 18.9\% and 33.2\% on the individual domains, above the baselines. Our RGB-D ensembled performance was 38.1\%, well above the Gaussian and Kr\"ahenb\"uhl et al.~\cite{krahenbuhl2015data} baselines. These results suggest that split-brain autoencoding is effective not just on \textit{Lab} images, but also on RGB-D data.

\section{Discussion}

We present split-brain autoencoders, a method for unsupervised pre-training on large-scale data. The split-brain autoencoder contains two disjoint sub-networks, which are trained as cross-channel encoders. Each sub-network is trained to predict one subset of raw data from another. We test the proposed method on \textit{Lab} images, and achieve state-of-the-art performance relative to previous self-supervised methods. We also demonstrate promising performance on RGB-D images. The proposed method solves some of the weaknesses of previous self-supervised methods. Specifically, the method (i) does not require a representational bottleneck for training, (ii) uses input dropout to help force abstraction in the representation, and (iii) is pre-trained on the full input data.

An interesting future direction of is exploring the concatenation of more than 2 cross-channel sub-networks. Given a fixed architecture size, e.g. AlexNet, dividing the network into $N$ disjoint sub-networks results in each sub-network becoming smaller, less expressive, and worse at its original task. To enable fair comparisons to previous large-scale representation learning methods, we focused on learning weights for a fixed AlexNet architecture. It would also be interesting to explore the regime of fixing the sub-network size and allowing the full network to grow with additional cross-channel encoders.

\section*{Acknowledgements}
\vspace{-1mm}

\noindent

We thank members of the Berkeley Artificial Intelligence Research Lab (BAIR), in particular Andrew Owens, for helpful discussions, as well as Saurabh Gupta for help with RGB-D experiments. This research was supported, in part, by Berkeley Deep Drive (BDD) sponsors, hardware donations by NVIDIA Corp and Algorithmia, an Intel research grant, NGA NURI, and NSF SMA-1514512. Thanks Obama.

\normalsize

\appendix

\section*{Appendix}

\noindent
In Section \ref{sec:app:analysis}, we provide additional analysis. In Section \ref{sec:app:implementation}, we provide implementation details.
\section{Additional analysis}
\label{sec:app:analysis}

\noindent
\textbf{Cross-Channel Encoder Aggregation Analysis} In Figure \ref{fig:ilsvrclin_comp}(c), we show variations on aggregated cross-channel encoders. To begin, we hypothesize that the performance improvement of split-brain autoencoders \textbf{Split-Brain Auto (cl,cl)} over single cross-channel encoders \textbf{L$\rightarrow$ab} is due to the merging of complementary signals, as each sub-network in \textbf{Split-Brain Auto} has been trained on different portions of the input space. However, the improvement could be simply due to an ensembling effect. To test this, we train a split-brain autoencoder, comprising of two \textbf{L$\rightarrow$ab} networks, \textbf{Ensemble L$\rightarrow$ab}. As seen in Figure \ref{fig:ilsvrclin_comp}(c) and Table \ref{tab:ilsvrclin}, the ensembled colorization network achieves lower performance than the split-brain autoencoder, suggesting that concatenating signals learned on complementary information is beneficial for representation learning.

We find that combining cross-channel encoders through concatenation is effective. We also test alternative aggregation techniques. As seen in Figure \ref{fig:ilsvrclin_comp}(c), training a single network to perform multiple cross-channel tasks \textbf{(L,ab)$\rightarrow$(ab,L)} is not effective for representation learning on full $Lab$ images. Adding in the autoencoder objective during training, \textbf{(L,ab,Lab)$\rightarrow$(ab,L,Lab)}, in fact lowers performance in higher layers.

Our proposed methods outperform these alternatives, which indicates that (i) our choice of aggregating complementary signals improves performance (ii) concatenation is an appropriate choice of combining cross-channel encoders.

\section{Implementation Details}
\label{sec:app:implementation}

\begin{table}[b]
\centering
\vspace{-.1in}
\begin{tabular}{C{10mm}C{5mm}C{5mm}C{5mm}C{5mm}C{5mm}C{5mm}}
\specialrule{.1em}{.1em}{.1em}
\multicolumn{7}{c}{\textbf{Fully Convolutional AlexNet \cite{krizhevsky2012imagenet} Architecture}} \\
\specialrule{.1em}{.1em}{.1em}
\specialrule{.1em}{.1em}{.1em}
\textbf{Layer} & \textbf{X} & \textbf{C} & \textbf{K} & \textbf{S} & \textbf{D} & \textbf{P} \\ \hline
\textbf{data} & 180 & * & -- & -- & -- & -- \\
\textbf{conv1} & 45 & 96 & 11 & 4 & 1 & 5 \\
\textbf{pool1} & 23 & 96 & 3 & 2 & 1 & 1 \\
\textbf{conv2} & 23 & 256 & 5 & 1 & 1 & 2 \\
\textbf{pool2} & 12 & 256 & 3 & 2 & 1 & 1 \\
\textbf{conv3} & 12 & 384 & 3 & 1 & 1 & 1 \\
\textbf{conv4} & 12 & 384 & 3 & 1 & 1 & 1 \\
\textbf{conv5} & 12 & 256 & 3 & 1 & 1 & 1 \\
\textbf{pool5} & 12 & 256 & 3 & 1 & 1 & 1 \\
\textbf{fc6} & 12 & 4096 & 6 & 1 & 2 & 6 \\
\textbf{fc7} & 12 & 4096 & 1 & 1 & 1 & 0 \\
\textbf{fc8} & 12 & * & 1 & 1 & 1 & 0 \\
\specialrule{.1em}{.1em}{.1em}
\end{tabular}
\caption{\textbf{Fully Convolutional AlexNet architecture used for pre-training}. \textbf{X} spatial resolution of layer, \textbf{C} number of channels in layer; \textbf{K} \texttt{conv} or \texttt{pool} kernel size; \textbf{S} computation stride; \textbf{D} kernel dilation \cite{chen2016deeplab,yu2015multi}; \textbf{P} padding; * first and last layer channel sizes are dependent on the pre-text task, last layer is removed during transfer evaluation.}
\vspace{-.1in}
\label{tab:arch-fc}
\end{table}

\begin{table}[b]
\centering
\begin{tabular}{C{10mm}C{5mm}C{5mm}C{5mm}C{5mm}C{5mm}C{5mm}C{5mm}C{5mm}}
\specialrule{.1em}{.1em}{.1em}
\multicolumn{9}{c}{\textbf{AlexNet Classification \cite{krizhevsky2012imagenet} Architecture}} \\
\specialrule{.1em}{.1em}{.1em}
\specialrule{.1em}{.1em}{.1em}
\textbf{Layer} & \textbf{X} & $\textbf{X}_d$ & \textbf{C} & $\textbf{F}_d$ & \textbf{K} & \textbf{S} & \textbf{D} & \textbf{P} \\ \hline
\textbf{data} & 227 & -- & * & -- & -- & -- & -- & -- \\
\textbf{conv1} & 55 & 10 & 96 & 9600 & 11 & 4 & 1 & 0 \\
\textbf{pool1} & 27 & 10 & 96 & 9600 & 3 & 2 & 1 & 0 \\
\textbf{conv2} & 27 & 6 & 256 & 9216 & 5 & 1 & 1 & 2 \\
\textbf{pool2} & 13 & 6 & 256 & 9216 & 3 & 2 & 1 & 0 \\
\textbf{conv3} & 13 & 5 & 384 & 9600 & 3 & 1 & 1 & 1 \\
\textbf{conv4} & 13 & 5 & 384 & 9600 & 3 & 1 & 1 & 1 \\
\textbf{conv5} & 13 & 6 & 256 & 9216 & 3 & 1 & 1 & 1 \\
\textbf{pool5} & 6 & 6 & 256 & 9216 & 3 & 2 & 1 & 0 \\
\textbf{fc6} & 1 & -- & 4096 & -- & 6 & 1 & 1 & 0 \\
\textbf{fc7} & 1 & -- & 4096 & -- & 1 & 1 & 1 & 0 \\
\specialrule{.1em}{.1em}{.1em}
\end{tabular}
\caption{\textbf{AlexNet architecture used for feature evaluation}. \textbf{X} spatial resolution of layer, $\textbf{X}_d$ downsampled spatial resolution for feature evaluation, \textbf{C} number of channels in layer; $\textbf{F}_d=\textbf{X}_d^2\textbf{C}$ downsampled feature map size for feature evaluation (kept approximately constant throughout), \textbf{K} \texttt{conv} or \texttt{pool} kernel size; \textbf{S} computation stride; \textbf{D} kernel dilation \cite{chen2016deeplab,yu2015multi}; \textbf{P} padding; * first layer channel size is dependent on the pre-text task e.g., 3 for the split-brain autoencoder or 1 for the $L\rightarrow ab (cl)$ cross-channel encoder}
\vspace{-.1in}
\label{tab:arch-cl}
\end{table}

Here, we describe the pre-training and feature evaluation architectures. For pre-training, we use an AlexNet architecture \cite{krizhevsky2012imagenet}, trained fully convolutionally \cite{long2015fully}. The network is trained with 180$\times$180 images, cropped from $256\times 256$ resolution, and predicts values at a heavily downsampled 12$\times$12 resolution. One can add upsampling-convolutional layers or use \textit{a trous}~\cite{chen2016deeplab}/dilated~\cite{yu2015multi} convolutions to predict full resolution images at the expense of additional memory and run-time, but we found predicting at a lower resolution to be sufficient for representation learning. See Table \ref{tab:arch-fc} for feature map and parameter sizes during pre-training time. We remove \texttt{LRN} layers and add \texttt{BatchNorm} layers after every convolution layer. After pre-training, we remove \texttt{BatchNorm} layers by absorbing the parameters into the preceding \texttt{conv} layers. The pre-training network predicts a downsampled version of the desired output, which we found to be adequate for feature learning.

During feature evaluation time (such as the ImageNet \cite{krizhevsky2012imagenet}, Places \cite{zhou2014learning}, and PASCAL \cite{everingham2010pascal} tests), the parameters are copied into an AlexNet classification architecture, shown in Table \ref{tab:arch-cl}. During the linear classification tests, we downsample feature maps spatially, so that each layer has approximately the same number of features.

\textbf{Quantization procedure} Zhang et al.~\cite{zhang2016colorful} use a class-rebalancing term, to over-sample rare colors in the training set, and a soft-encoding scheme for $\mathcal{H}$. These choices were made from a graphics perspective, to produce more vibrant colorizations. In our classification colorization network, \textbf{L$\rightarrow$ab(cl)}, our objective is more straightforward, as we do not use class-rebalancing. In addition, we use a 1-hot encoding representation of classes, rather than soft-encoding. The simplification in the objective function achieves higher performance on ImageNet and Places classification, as shown on Tables \ref{tab:ilsvrclin} and \ref{tab:placeslin}.
\section{Change Log}

\noindent
\textbf{v1} Initial Release.

\noindent
\textbf{v2} Paper accepted to CVPR 2017. Updated Table \ref{tab:pascal} with results for Misra et al. \cite{misra2016shuffle} and Donahue et al. \cite{donahue2016adversarial} with $112\times 112$ resolution model. Updated Table \ref{tab:ilsvrclin}, rows \textbf{L$\rightarrow$ab (cl)} and Zhang et al. \cite{zhang2016colorful} with corrected values. Supplemental material added.

\noindent
\textbf{v3} CVPR 2017 Camera Ready. Added references to concurrent work \cite{pathak2017learning,larsson2017colorization}. Various changes to text.

{\small
\bibliographystyle{ieee}
\bibliography{egbib}
}

\end{document}